\documentclass[runningheads]{llncs}
\usepackage{graphicx}
\usepackage[a4paper, total={6in, 7in}]{geometry}
\usepackage{csquotes}
\usepackage{multirow}

\usepackage{booktabs}
\usepackage{graphicx}

\usepackage{float}
\usepackage{multirow}
\usepackage{makecell}

\begin{document}

\title{Unsupervised Driver Behavior Profiling leveraging Recurrent Neural Networks}

%

\newcommand*\samethanks[1][\value{footnote}]{\footnotemark[#1]}

\author{
Young Ah Choi\inst{1} \thanks{Equal Contribution} \and
Kyung Ho Park\inst{2} \samethanks[1] \and
Eunji Park\inst{1} \and
Huy Kang Kim\inst{1} \thanks{Corresponding author}
}

\authorrunning{Choi et al.}
%

\institute{
School of Cybersecurity, Korea University, Republic of Korea
\email{\{choiya3168,epark911,cenda\}@korea.ac.kr} \and
SOCAR, Republic of Korea\\
\email{kp@socar.kr}
}

\maketitle  

\begin{abstract}
In the era of intelligent transportation, driver behavior profiling has become a beneficial technology as it provides knowledge regarding the driver's aggressiveness. Previous approaches achieved promising driver behavior profiling performance through establishing statistical heuristics rules or supervised learning-based models. Still, there exist limits that the practitioner should prepare a labeled dataset, and prior approaches could not classify aggressive behaviors which are not known a priori. In pursuit of improving the aforementioned drawbacks, we propose a novel approach to driver behavior profiling leveraging an unsupervised learning paradigm. First, we cast the driver behavior profiling problem as anomaly detection. Second, we established recurrent neural networks that predict the next feature vector given a sequence of feature vectors. We trained the model with normal driver data only. As a result, our model yields high regression error given a sequence of aggressive driver behavior and low error given at a sequence of normal driver behavior. We figured this difference of error between normal and aggressive driver behavior can be an adequate flag for driver behavior profiling and accomplished a precise performance in experiments. Lastly, we further analyzed the optimal level of sequence length for identifying each aggressive driver behavior. We expect the proposed approach to be a useful baseline for unsupervised driver behavior profiling and contribute to the efficient, intelligent transportation ecosystem.
\keywords{Driver Behavior Profiling \and Unsupervised Learning \and Recurrent Neural Networks}
\end{abstract}

\section{Introduction}
Recently, driver behavior profiling has become a significant technology in the era of intelligent transportation. Driver behavior profiling implies a sequence of operations that collects driving data, analyzes the driving pattern, and provides appropriate actions to the driver to achieve the benefit of safety and energy-aware driving \cite{eren2012estimating}. For example, if the driver behavior profiling can detect sudden driving pattern changes (in the cases that aggressive driving or car-theft), it can contribute to the road-safety highly. To collect the data from the vehicle, prior works leveraged various sensors such as a telematics system or Controller Area Network (CAN) embedded in a car. As drivers in these days always carry their smartphones to the car, several studies utilized sensor measurements extracted from mobile devices, analyzed driver behaviors, and commercialized industrial applications such as Pay-How-You-Drive \cite{carvalho2017exploiting}. Early approaches to driver behavior profiling started with statistical analyses. Previous studies scrutinized the driving data and figured our particular heuristic rules to identify aggressive driver behaviors. While statistical approaches were easy to implement and deploy in the wild, there exist several limits. First, the practitioners should establish detection rules on each aggressive driver behavior, and it creates a particular amount of resource consumption in the real world. Second, heuristic rules necessitated frequent updates if the pattern of driver behavior changes \cite{choi2007analysis}, \cite{hu2017abnormal}, \cite{zhang2017safedrive}.

Along with the development of machine learning algorithms, recent approaches started to resolve the aforementioned limits by training a large amount of driving data to the learning-based models. As machine learning algorithms could effectively analyze the unique characteristics of aggressive driver behaviors, prior learning-based methods achieved an improved driver behavior profiling performance. Nevertheless, there exist several hurdles to deploy learning-based driver behavior profiling approaches. As learning-based approaches trained the model under the supervised learning paradigm, the practitioner must prepare a finely-labeled dataset, which consumes a particular amount of resource consumption in the real world. Furthermore,  the trained model could only identify aggressive driver behaviors known a priori. If an unseen aggressive driver behavior happens in the wild, the model trained under the supervised learning paradigm could not classify it as aggressive driver behavior \cite{carvalho2017exploiting}, \cite{martinez2016driver}, \cite{zhang2019deep}. In pursuit of resolving these limits of the supervised learning paradigm, several works cast the driver behavior profiling task as an anomaly detection paradigm. They provided normal driving data only to the model and let the model identify any other patterns distinct from the trained normal patterns as anomalies (\textit{i.e.}, aggressive driver behaviors). Previous studies employed an unsupervised learning paradigm to scrutinize the normal driving data and achieved a promising driver behavior profiling performance.

\begin{figure}[htb!]
\centering
\begin{tabular} {c} \\
\includegraphics[width=0.8\textwidth]{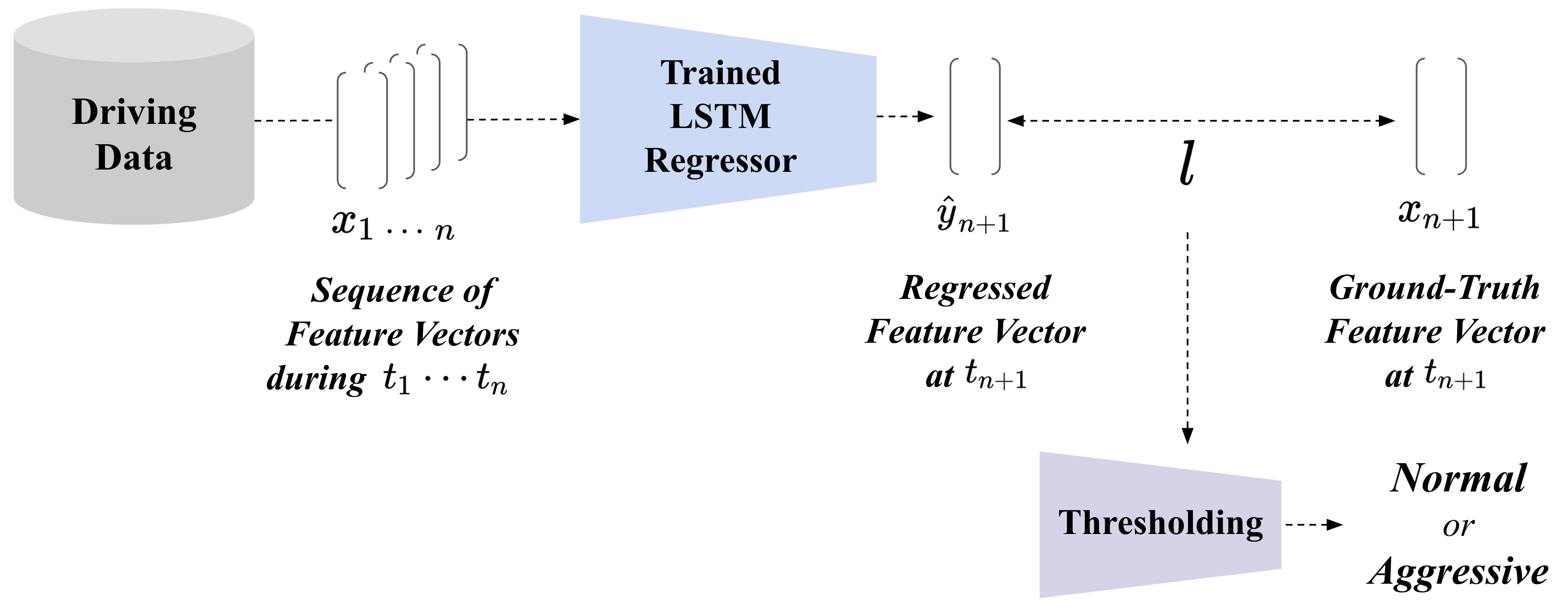} \cr
\end{tabular}
\caption{The architecture of the proposed driver behavior profiling approach}
\label{fig:lstm}
\end{figure}

Following the motivation of the aforementioned unsupervised paradigm, our study proposes a novel approach to driver behavior profiling that only requires normal driving data during the model training. An overview of our approach is illustrated in Fig. \ref{fig:lstm}. First, our approach establishes a sequence of feature vectors from the log-level driving data. Second, we trained the model to regress the feature vector right after the sequence given a sequence of feature vectors. Suppose that we set a single sequence consisting of 10 feature vectors recorded from the timestamp 1 to 10. In this case, we designed the model to predict the feature vector at timestamp 11, given the sequence of feature vectors. After the model analyzes a particular amount of normal driving data, we expected it would learn the pattern of normal driver behaviors during the training phase. Note that we designed the model with recurrent neural networks to capture the pattern of time-series characteristics of driving data. Third, we measured the error between the predicted feature vector and the ground-truth feature vector. If the given sequence was originated from normal driver behaviors, the error would be small as the trained model already learned the normal driver behaviors' patterns. On the other hand, the error would be large if the given sequence illustrates aggressive driver behavior. As aggressive driver behaviors' characteristics are not trained during the training phase, the trained model presumably fails to predict the feature vectors of aggressive driver behaviors. Lastly, we analyzed this error as an adequate flag to identify aggressive driver behaviors from normal driver behaviors; therefore, we classified a particular sequence as an anomaly (\textit{i.e.}, aggressive driver behavior) if the error goes larger than a particular threshold level. 

The contributions of our study are as follows:
\begin{itemize}
    \item We designed a novel approach to driver behavior profiling under the unsupervised learning paradigm.
    
    \item We examined our approach effectively identifies 6 types of aggressive driver behavior types unless the model is only trained with normal driving data. 
    
    \item We experimentally discovered the optimal size of sequence length varies along with aggressive driver behavior types.
    
\end{itemize}

\section{Literature Review}
In this section, we briefly scrutinized prior studies on driver behavior profiling in three categories: statistical approaches, supervised approaches, and unsupervised approaches.

\subsection{Statistical Approaches}
The statistical approach includes driving data analysis and statistic-based driving events classification. Berndt \textit{et al.} conducted a study to infer the driver's intention for lane change and rotation. With data from internal sensors, the feature signals got generated for identifying the driver's intention. Through pattern matching using Hidden Markov Models (HMMs), they proposed a method to determine whether upcoming action is dangerous and provide feedback \cite{berndt2009driver}. Choi \textit{et al.} conducted the study with three objectives: classifying the actions of driving, detecting distractions, and identifying a driver. First, they used HMMs to classify actions and detect distractions by capturing dynamic driving features. Subsequently, they identified drivers while modeling features by a driver with the Gaussian Mixture Model (GMM) \cite{choi2007analysis}. SafeDrive, a proposed system by Zhang \textit{et al.}, performed real-time detection of abnormal drivings. It created state graphs on the behavior model of driving data and compared them to the online behavior streams. If the comparison showed a significant difference, they classified the case as abnormal driving \cite{zhang2017safedrive}. Eren \textit{et al.} proposed a methodology that analyzed driving behavior and informed how to reduce fuel consumption. They established three modules: an action detecting module with a linear model, a fuzzy module for evaluating fuel consumption, a proposing module. The last one suggested more efficient driving patterns under eight rules \cite{eren2012estimating}. Dai \textit{et al.} focused on detecting dangerous driving related to drunk drivers. They used data from the horizontal accelerometer using sensors from smartphones. The predefined pattern of drunk driving went through multiple matching rounds with collected data for efficient detection \cite{dai2010mobile}. The studies above achieved reliable performance of classification. Most of them solved the problem in a way that used predefined rules or patterns. However, they have a limitation that the statistical approach could not detect abnormal driving of unknown types. 

\subsection{Supervised Approaches}
Approaches using supervised learning models have been studied to detect more diverse driving behaviors. Zhang \textit{et al.} researched driving behavior classification through high-dimensional data from multi-sensors. They extracted features using a convolutional neural network (CNN) after modeling features by exploring correlations between data points. The classification model was designed with two types of Attention-based recurrent neural networks (RNN) \cite{zhang2019deep}. Amata \textit{et al.} proposed a prediction model, depending on two factors: the driving styles and traffic conditions. For action prediction with only driving styles, they constructed a linear regression model. They also used a Bayesian network to predict if the driver would decelerate in a given situation considering both factors \cite{amata2009prediction}. Similarly, Olaviyi \textit{et al.} also conducted a study to predict driving behavior. A system using a deep bidirectional recurrent neural network (DBRNN) recognized traffic conditions and current driving styles and then predicted driving actions ahead \cite{olabiyi2017driver}. Wu \textit{et al.} established learning models for classifying driving behaviors such as accelerating, braking, and turning, rather than just identifying normal/abnormal drivings. Their study went through a data processing process that separated segments based on a time window and managed models using various machine learning techniques such as support vector machine (SVM), logistic regression, and K-nearest neighbors (KNN) \cite{wu2016novel}. Chen \textit{et al.} proposed a system called Driving Behavior Detection and iDentification System (3D). Similar to \cite{wu2016novel}, they classified abnormal drivings as six behaviors. The proposed system classified incoming data streams in real-time, using SVM \cite{chen2015d}. These methods surpassed the boundaries of statistical methods and showed high detection performance as a result of research. Most of the data used in these studies were collected under experimental environments; thus, the label they had were accurate. However, in reality, the accessibility to unlabeled data is much higher, which can be seen as a limitation of the supervised approach.

\subsection{Unsupervised Approaches}
The prior studies identified driving behaviors and styles with unlabeled data for better application to the real world. Fugiglando \textit{et al.} researched to classify multiple drivers by driving styles using unlabeled data. The features related to drivers' similarities were selected to cluster them with similar driving styles. Then they successfully classified drivers by proceeding to K-means clustering \cite{fugiglando2018driving}. Van Ly \textit{et al.} conducted a similar study and further sought to provide feedback based on driving styles. They first utilized K-means clustering to profile the drivers' styles using collected data such as accelerating, braking, and turning from inertial sensors. After the profiling, the feedback-providing process was implemented with a model using SVM \cite{van2013driver}. Mitrovic \textit{et al.} worked on driving style classification and prediction of driving actions. For short-term prediction, they trained a multilayer perceptron (MLP) model using 10 minutes of data. Moreover, they proposed long-term prediction through K-means clustering, and it used data with more divergence of driving actions than the short-term prediction \cite{mitrovic2001machine}. Mantouka \textit{et al.} studied to detect unsafe driving styles separately with an unsupervised approach. The proposed methodology had a two-stage clustering architecture, in which the first stage was to detect unsafe driving, and the second stage classified unsafe driving into six classes \cite{mantouka2019identifying}.

\section{Proposed Methodology}
\subsection{Dataset Acquisition} 
In this study, we utilized a publicly-distributed dataset illustrate in \cite{carvalho2017exploiting} and \cite{ferreira2017driver}. The dataset includes driving data of two experienced drivers who drove the car for more than 15 years. The drivers drove the paved road on a sunny day. The IMU measurements embedded in the smartphone are established while two drivers drove the route with Honda Civic four times during near 13 minutes. To minimize the influence from the hardware installation, the smartphone was fixed to the vehicle's windshield and never moved or manipulated while data was being accumulated. The features of the collected driving data are composed of acceleration, linear acceleration, magnetometer, and gyroscope. Each feature includes three values at the axis $x$, $y$, $z$, the total number of features goes 12. Note the detailed explanation of the data collection environment is illustrated in \cite{carvalho2017exploiting} and \cite{ferreira2017driver}. The dataset includes seven types of driver behaviors: 1 normal driver behavior and six aggressive driver behaviors of aggressive brake, aggressive acceleration, aggressive left turn, aggressive right turn, aggressive lane change to the right, and aggressive lane change to the left.

\subsection{Feature Engineering}
After we acquired the dataset, we extracted feature vectors from the raw log-level driving data. We aim to transform raw driving data into feature vectors and establish pairs of (sequence of feature vectors, feature vector right after the sequence), which becomes an input and output of the model, respectively. The feature engineering process consists of three steps: timestamp calibration, scaling, and window sliding.

\subsubsection{Timestamp Calibration}
The acceleration, linear acceleration, magnetometer, and gyroscope sensor measurements are recorded in different frequencies in raw data. Since each feature's number of data points is different for a particular period, it is necessary to apply a frequency calibration to create a fixed shape of the feature vector. We performed the frequency calibration process by designing a scheme that downsampled the data of features with high frequency and upsampled features with low frequency. First and foremost, following the prior approaches illustrated in \cite{ferreira2017driver} and \cite{carvalho2017exploiting}, we set the target frequency to 50Hz and integrated the frequencies of all features to 50Hz. For features with a frequency higher than 50Hz, the first value during an initial period is selected as a representative value to be downsampled for the sake of computation convenience. Note that we experimentally verified that the index of representative values did not show any meaningful difference in driver behavior learning patterns. Upsampling uses a zero-order-hole scheme to extract an approximation from the first value of the initial period in feature data with a frequency less than the target frequency. Zero-order-hold is a scheme commonly used in signal processing, and it creates continuous data by duplicating constant values approximating values between them from given data.
In a nutshell, the upsampling and downsampling approach is chosen following the target frequency (50Hz); therefore, we could have established fixed-shape feature vectors.

\subsubsection{Scaling}
After we resolved an obstacle of frequency differences, we should have to manage the scale differences among features. We mitigated this difference by applying a MinMax scaler as we expected different scales of each feature to hinder the model training. We established a MinMax scaler following the definition described in Eq. \ref{eu_eqn}. 
\begin{equation} \label{eu_eqn}
X_{scaling} = \frac{X_{i} - X_{min}}{X_{max} - X_{min}}.
\end{equation}
We established a MinMax scaler, since our approach only utilizes normal driving data during the model training. Note the scaler established from normal driving data is also utilized during the inference stage. Therefore, we could have unified the scale of each feature under the particular range by applying the scaler.

\begin{figure}[htb!]
\centering
\begin{tabular} {c} \\
\includegraphics[width=0.65\textwidth]{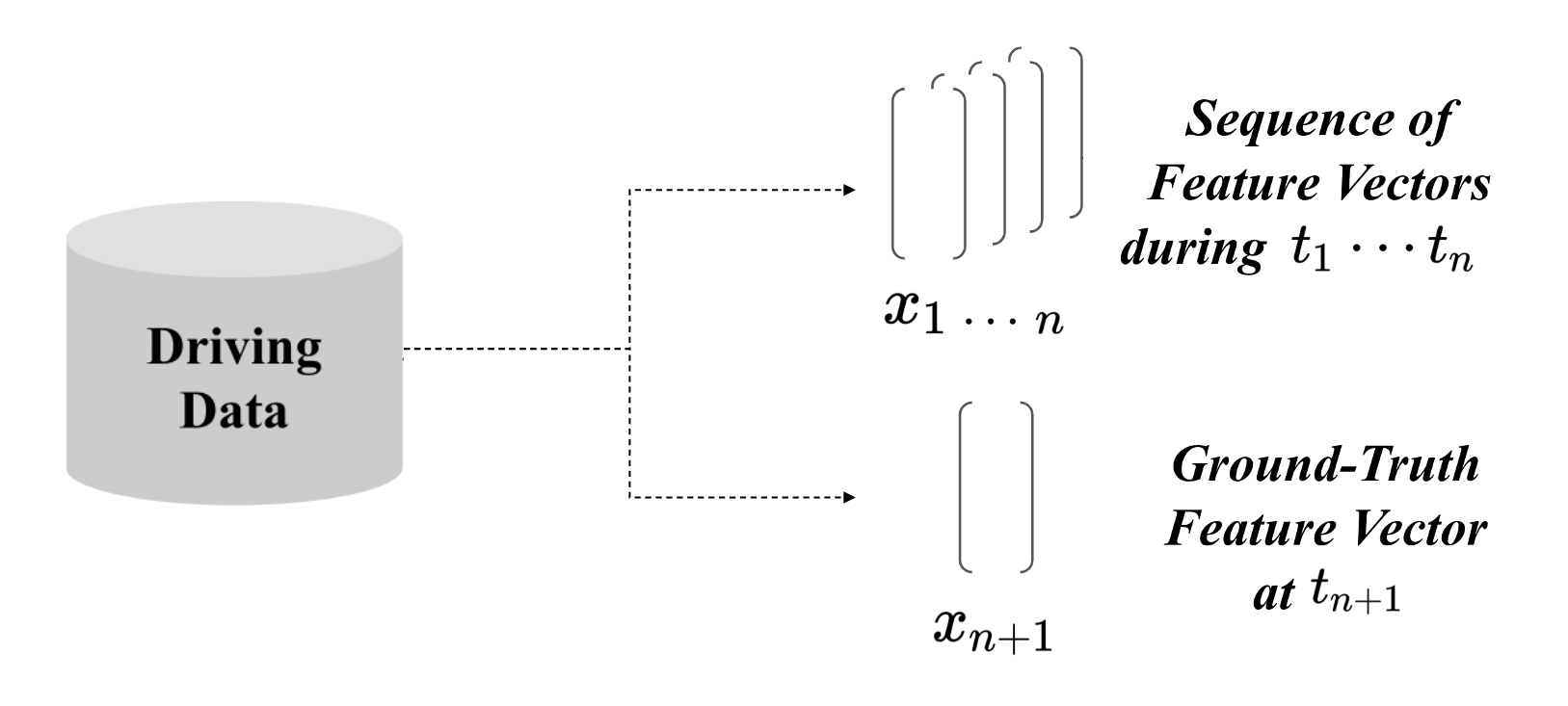} \cr
\end{tabular}
\caption{Pairs of data consists of the feature vector sequences and the ground-truth feature vector}
\label{fig:pair}
\end{figure}

\subsubsection{Window Sliding}
We set a particular window size and slid along with the timestamps to generate a pair of data consists of (feature vector sequence, single feature vector right after the sequence). The aforementioned pair becomes an input to the model and the ground truth for the model's prediction, respectively. Suppose that we set the window size as 25. In this case, the shape of the feature vector sequence (input) becomes (25, 12) as there exist 12 feature types, and the ground truth value's shape becomes (1, 25). Therefore, we established the dataset composed of feature vector sequences and its target ground truth. The composition is illustrated in Fig. \ref{fig:pair}. Note that we fully acknowledge different window sizes would influence the effectiveness of learning the pattern of driving data; thus, we processed the window sliding with window sizes of 25, 50, 100, and 200 and examined the driver behavior profiling performance in the later section.

\subsection{Training Stage}
During the training stage, we trained the model to learn unique characteristics of the normal driving data. As prior researches analyzed in \cite{zhang2019deep}, we expected the driving data to have temporal dynamics in feature space; therefore, we employed recurrent form of neural networks to scrutinize the time-series pattern of driving data effectively. Among various forms of recurrent neural networks, we utilized Long Short Term Memory (LSTM) neural networks as a model. 
We implemented the LSTM layers and added fully connected layers at the end of recurrent layers to regress the target value.
We designed the model to sequence feature vectors as input and predicted the feature vector recorded right after the given input sequence. We set the earning objective to minimize the difference (error) between the predicted feature vector and the ground truth by employing the loss function as Mean Squared Error (MSE). Moreover, we utilized $l1$ and $l2$ regularizers to evade the risk of overfitting and optimized the model with Adam optimizer. 

\subsection{Inference Stage}
During the inference stage, the key takeaway is classifying whether a given sequence of feature vectors illustrates aggressive driver behaviors or normal driver behaviors. Suppose the given sequence of feature vectors represents a normal driver behavior. In this case, the regression loss during the inference stage would be small as the pattern of normal driver behaviors is already trained during the training stage. On the other hand, the regression loss would become large when the given sequence implies aggressive driver behaviors. As the characteristics of aggressive driver behaviors are not trained a priori, the model would not be able to precisely regress the target feature vector of aggressive driver behaviors. Following the aforementioned analogies, we provided validation sequences into the trained model, calculate the loss between the predicted value and the target value, and classified the sequence as an aggressive driver behavior if the loss goes larger than a particular threshold level. Note that we acknowledged the threshold level can influence the classification performance; thus, we employed adequate evaluation metrics to validate our approach's effectiveness, which will be further elaborated in the following section.

\section{Experiments}
\subsection{Experiment Setup}

During the training stage, we sampled a particular number of normal driving data to train the model. During the inference stage, we utilized both normal driving data and 6 types of driving data to perform a binary classification; therefore, we conducted 6 types of test sets to measure our approach's effectiveness: aggressive right turn, left turn, right lane change, left lane change, brake, and acceleration. Throughout experiments, we employed the evaluation metric as Receiver Operating Characteristic - Area Under Curve (ROC-AUC). As our approach's driver behavior profiling performance varies along with threshold levels, we aimed to measure overall effectiveness at numerous thresholds; thus, ROC-AUC was an adequate metric to validate the performance. Suppose the difference between regression losses from normal driving data and aggressive driving data is large. In this case, the ROC-AUC would go large, and it implies our approach successfully discriminated normal driver behaviors and aggressive driver behaviors. Experiment results are shown in Table \ref{experiment result}.

\subsection{Experiment Result}

\begin{table}\label{result}
    \resizebox{1.0\columnwidth}{!}{
    \begin{minipage}{\textwidth}
    \renewcommand{\arraystretch}{1.3}
    \centering
    \caption{{Experiment result on our approach. Our approach precisely achieved driver behavior profiling performance in general but failed to identify aggressive acceleration.}}\label{experiment result}
    \begin{tabular}{ccccccc}
        \toprule
            \multicolumn{2}{c}{\multirow{2}{*}{}} & \multicolumn{4}{c}{window} & \multirow{2}{*}{\begin{tabular}[c]{@{}c@{}} Avg of AUC\\ by label \end{tabular}} \\ \cmidrule(lr){3-6}
            \multicolumn{2}{c}{} & \multicolumn{1}{c}{200} & \multicolumn{1}{c}{100} & \multicolumn{1}{c}{50} & \multicolumn{1}{c}{25} & \\ \hline \hline
        \multicolumn{1}{c}{\multirow{6}{*}{Label}} & 
        \multicolumn{1}{c}{\begin{tabular}[c]{@{}c@{}}Aggressive Right Turn\end{tabular}} &
          0.9648 &
          0.9722 &
          \textbf{0.9725} &
          0.9647 &
          0.9686 \\ \cmidrule(l){2-7} 
        \multicolumn{1}{c}{} &
          \multicolumn{1}{c}{\begin{tabular}[c]{@{}c@{}}Aggressive Left Turn\end{tabular}} &
          0.9213 &
          0.9409 &
          \textbf{0.9421} &
          0.9202 &
          0.9311 \\ \cmidrule(l){2-7} 
        \multicolumn{1}{c}{} &
          \multicolumn{1}{c}{\begin{tabular}[c]{@{}c@{}}Aggressive Right Lane Change\end{tabular}} &
          \textbf{0.9028} &
          0.8872 &
          0.8979 &
          0.8691 &
          0.8892 \\ \cmidrule(l){2-7} 
        \multicolumn{1}{c}{} &
          \multicolumn{1}{c}{\begin{tabular}[c]{@{}c@{}}Aggressive Left Lane Change\end{tabular}} &
          0.8962 &
          0.8855 &
          \textbf{0.9026} &
          0.8747 &
          0.8897 \\ \cmidrule(l){2-7} 
        \multicolumn{1}{c}{} &
          \multicolumn{1}{c}{\begin{tabular}[c]{@{}c@{}}Aggressive Brake\end{tabular}} &
          \textbf{0.9057} &
          0.8889 &
          0.8954 &
          0.8936 &
          0.8959 \\ \cmidrule(l){2-7} 
        \multicolumn{1}{c}{} &
          \multicolumn{1}{c}{\begin{tabular}[c]{@{}c@{}}Aggressive Acceleration\end{tabular}} &
          \textbf{0.7326} &
          0.7012 &
          0.6769 &
          0.7241 &
          0.7087 \\ \hline \hline
        \multicolumn{2}{c}{\begin{tabular}[c]{@{}c@{}}Average of AUC by window\end{tabular}} &
          \multicolumn{1}{c}{\textbf{0.8872}} &
          \multicolumn{1}{c}{0.8793} &
          \multicolumn{1}{c}{0.8812} &
          \multicolumn{1}{c}{0.8744} &
          \\ \bottomrule
        \multicolumn{2}{c}{Average of AUC} &
          \multicolumn{1}{c}{} &
          \multicolumn{1}{c}{} &
          \multicolumn{1}{c}{} &
          \multicolumn{1}{c}{} &
          \textbf{0.8805} \\ \bottomrule \bottomrule
    \end{tabular}
    \end{minipage}}
\end{table}

\noindent
Experimental results can be checked by window size and aggressive driving pattern type.
Among the four windows in Table \ref{experiment result}, the models with a window size of 200 showed the highest performance (0.89), while the model with a window size is 100 or 50 showed 0.88, and the 25 window model showed a slight difference with 0.87.

By aggressive driving pattern type, \textbf{Aggressive Turn} showed the best detection rate with an average value of 0.97 and 0.93. The \textbf{Aggressive Brake} showed the result of 0.9, slightly better than the \textbf{Aggressive Lane Change} of 0.89. Since the degree of variation in the lane change was smaller than that of right and left turns; there was also a slight difference in the detection rate.

In the case of \textbf{Aggressive Acceleration}, it is difficult to distinguish it from the normal driver. In fact, this label showed significantly less difference than the other aggressive labels. Five other aggressive drivings show the high precision with 0.8 to 0.9; However, in the case of the \textbf{Aggressive Acceleration}, all windows showed low performance with about 0.677 to 0.724.

For other labels, the inertia value measured with IMU sensors was used as a feature. This feature is useful to detect significant value changes when drivers do right turn, left turn, brake, and lane change. 
That means this feature is sensitive to detect the aggressive driving pattern. Likewise, our model showed a good performance for \textbf{Aggressive right and left turns}, \textbf{Aggressive brake}, and \textbf{Aggressive lane change}.
  


\textbf{Aggressive Acceleration and Brake} representative events that provide plentiful information about the aggressive driving pattern. For \textbf{Aggressive Acceleration and Brake}, the window size of 200 showed a good performance.

In case of the \textbf{Aggressive turn and lane change}, the window size of 50 showed a good performance in the experiment. 

To summarize, our proposed model trained with only normal driving patterns can successfully detect aggressive driving patterns. Also, we present the window sizes which show a good performance for each labeled event as shown in the experiment result.  

\section{Conclusions}
Throughout the study, we propose a novel approach to driver behavior profiling by leveraging recurrent neural networks under an unsupervised learning paradigm. First, we extracted fixed-shape feature vectors from raw log-level smartphone sensor measurements. Second, we designed LSTM regressors that predict the next feature vectors given a sequence of feature vectors and trained the model with normal driving data only to learn time-series characteristics of normal driver behaviors. During the inference stage, we figured out the trained model results in high regression error given a sequence of feature vectors derived from aggressive driver behaviors. As the trained model only understands the pattern of normal driver behaviors, it fails to precisely predict the next feature vector followed by the given sequence. On the other hand, the model provides low regression error given normal sequences as their patterns are already trained during the training stage. By thresholding the aforementioned regression error, our approach can effectively classify aggressive driver behaviors from normal behaviors. 
Although we showed a concrete baseline of unsupervised driver behavior profiling, we fully acknowledge there should be further improvements and strict validations on our work for real-world deployment. We shall perform further external validations of our approach with various drivers, different types of cars, different routes, and several types of smartphones. 
Along with the aforementioned validations and experiments, we expect the proposed driver behavior profiling approach can provide various benefits to society.

\section{Acknowledgement}
This work was supported by Institute of Information \& communications Technology Planning \& Evaluation (IITP) grant funded by the Korea government (MSIT) (No.2021-0-00624, Development of Intelligence Cyber Attack and Defense Analysis Framework for Increasing Security Level of C-ITS).


\end{document}